# Absolute distance prediction based on deep learning object detection and monocular depth estimation models

Armin MASOUMIAN[a,1], David G. F. MAREI[a], Saddam ABDULWAHAB[a], Julián CRISTIANO[a], Domenec PUIG[a] and Hatem A. RASHWAN[a]

[a]*DEIM, Rovira i Virgili University, 43007 Tarragona, Spain*

**Abstract.** Determining the distance between the objects in a scene and the camera sensor from 2D images is feasible by estimating depth images using stereo cameras or 3D cameras. The outcome of depth estimation is relative distances that can be used to calculate absolute distances to be applicable in reality. However, distance estimation is very challenging using 2D monocular cameras. This paper presents a deep learning framework that consists of two deep networks for depth estimation and object detection using a single image. Firstly, objects in the scene are detected and localized using the You Only Look Once (YOLOv5) network. In parallel, the estimated depth image is computed using a deep autoencoder network to detect the relative distances. The proposed object detection based YOLO was trained using a supervised learning technique, in turn, the network of depth estimation was self-supervised training. The presented distance estimation framework was evaluated on real images of outdoor scenes. The achieved results show that the proposed framework is promising and it yields an accuracy of 96% with RMSE of 0.203 of the correct absolute distance.

## 1. Introduction

For enabling fully autonomous driving and navigation, one of the main challenges is to achieve reliable and accurate obstacles detection. Many works have been proposed to cope with the problem of obstacles detection [1]. Object detection and distance prediction are effectively used in a variety of different fields such as industrial robots [2], robots for research [3], self-driving cars [4] and etc. Regarding object detection, to successfully navigate the environment, the moving system must have knowledge about the objects in its immediate vicinity. Among many sensors available for object detection (e.g. LIDAR sensors) we are primarily interested in a camera-based vision for indoor/outdoor navigation. Thus, object recognition based cameras refers to a collection of related tasks for identifying objects in digital photographs.

With the progress of deep learning networks (e.g., Convolutional Neural Networks (CNN)), many accurate methods for object recognition have been developed. For instance, region-Based CNN, or R-CNNs [5], are a family of techniques for addressing object localization and recognition tasks, designed for model performance. In turn, You

---

[1]Corresponding Author: E-mail: masoumian.armin@gmail.com

Only Look Once, or YOLO [6], is a second family of techniques for object recognition designed for a fast responseand real-time applications.

Region-based detectors include two stages. Firstly, the model suggests a set of regions of interests (ROIs) by a regional proposal network. Since the potential bounding box candidates can be infinite, the proposed regions are sparse. Secondly, the region candidates are then processed by a classifier. In turn, the one-stage family skips the region proposal stage and it directly runs the detection over a dense sampling of possible locations. This yields that the one-stage detectors are faster and simpler, but might potentially reduce the performance a bit. Since YOLO has the advantage of being much faster than other networks in the one-stage family. Besides it achieved comparable results to the state of the art and still maintains accuracy. The predictions depend on the global context of the input image. Consequently, our proposed framework will be based on the YOLO architecture as a baseline.

Regarding distance prediction, it is important to estimate depth maps from the input images. For depth estimation, most computer-vision systems depend on stereo vision by following several time-consuming stages, such as unipolar geometry, rectification and matching. Alternatively, when stereo vision is not useful or applicable, LIDAR cameras can be used for many applications for mobile robots. However, LIDAR sensors are very costly, and most depth cameras have serious limitations in real environments, such as the synchronization of the optical and imaging elements [7]. With the deep learning spread, many works have been proposed for monocular depth estimation that is the task of estimating scene depth using a single image. The appearance of objects significantly changes with their pose. Estimating a depth map from a 2D image is an important step in order to determine the 3D pose of the objects present in a scene. Monocular depth estimation based on deep learning methods can be performed by supervised [8] or unsupervised [9] learning techniques. Supervised methods perform better accuracy, however the depth maps of images is needed for training which is difficult to get in real scenarios. On the other hand, unsupervised methods do not require original depth maps, thus, the performance is degraded a bit.

Thus, in this paper, we propose a new framework to predict the absolute distance of each object in 2D images captured from a monocular camera, based on estimating depth images using self-supervised deep learning and supervised deep learning object detection. The contributions of this paper are:

- Developing a deep object detection model based on two-stage YOLOv5 architecture. A lightweight model that can be easily deployed on embedded systems and devices with a limited memory and CPU.
- Developing an unsupervised depth and pose estimation deep learning model based on an autoencoder network.
- Integrating the two models in a framework for absolute distance estimation of obstacles. Integrating the two models will not affect the overall efficiency of the proposed models, because the two models are structurally independent, and the whole framework is executed by multi processes, meaning that each model has a separate process responsible for it.

Section 2 introduces previous background review, section 3 explains our proposed methodology and section 4 describes our performed experiments.

## 2. Background Review

In this section, we summarize the state-of-the-art for both systems: depth estimation and object detection.

*2.1. Object Detection*

Object detection is a computer vision technique that allows a system to locate and identify an object in an image or video and detecting a bounding box around each one of them. It is one of the most challenging issues in the field of computer vision as the object detection model is trained to identify objects within a dataset and it cannot identify an object that is not labeled during the training and this is considered as one of the limitations. However, trained object detection models can always be retrained again to obtain new knowledge about new objects. Object detection techniques are used in applications like self-driving cars, video surveillance or crowd counting. There are some popular object detection algorithms like YOLO [6], R-CNN [5] and MobileNet [10]. In this paper the YOLO Algorithm has been chosen for object detection. As it is considered as the state-of-the-art right now and it produced the needed results in the testing phase.

The YOLO object detection model has several different versions through the years. The YOLOv1 paper was published in 2015 and later the subsequent versions were published the next years until it reached YOLOv5 in 2020 and it is considered as the state-of-the-art due to its good performance and efficiency and constantly being improved.

*2.2. Depth Prediction*

Depth and ego motion estimation is one of the fundamental challenges in computer vision. Two different methods exist to achieve these goals, supervised deep learning and unsupervised deep learning models.

*2.2.1. Supervised Depth Estimation*

Predicting a depth from a single image is an innately difficult task as the same image can project multiple conceivable depths. To prevent this, predicted depth needed to have some relationship with color image. There are various approaches such as end to end [11], sense sampling of non- parametric [12], optical flow [13], transfer learning [14] and combining local predictions [15], have been done. In supervised methods, the original depth maps of images are used to train alongside color images. This helps the system to learn better and therefore the results of supervised methods usually have better performance than unsupervised methods. However in reality it is difficult to construct the depth maps in real time, and to do that, the use of a stereo camera or 3D LIDAR is necessary.

*2.2.2. Unsupervised Depth Estimation*

To avoid the aforementioned problems, unsupervised methods have been used for training the systems, only original images and pre trained models such as DenseNet

[16], ResNet [17], and ImageNet [18] are needed. Regarding unsupervised methods, various approaches for depth estimation have been proposed, such as generative adversarial networks [19], temporal information [20] and separate pose networks [21].

## 3. Method

In this work, as shown in Figure 1, we used two parallel deep networks: one for object detection and the other for depth estimation. The predicted depth is extracted from DepthNet. In turn with YOLOv5, the objects inside the image is localized and classified. Furthermore, the localization of each object defined by bounded boxes is detected on the estimated depth image. Finally, the relevant distance of an object is calculated by the median estimated distance of all pixels inside the defined bounded box.

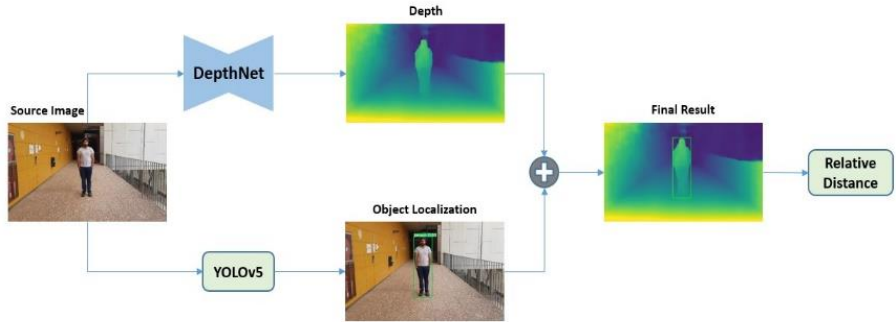

**Figure 1.** An illustration of the overall framework

Regarding DepthNet, it has two networks: one for estimating depth images and the other for estimating the image pose. Both networks are based on autoencoder networks that consist of two serial networks: encoders and decoders. For encoder of depth and pose networks, the ResNet pre-trained weight has been used for extracting the features and representing the input images. The ResNet-50 has been used as a backbone network for our depth prediction, while ResNet-18 has been used for pose estimation. In both networks, the first step before entering the first layer of the ResNet is a block — called here Conv1 — consisting on a convolution + batch normalization + max pooling operation. Then, four blocks of the ResNet are used. Regarding the decoder, each layer consists of a deconvolutional and upsampling, as shown in Figure 2. The last layer of decoder is the estimated depth map. Besides, we used our initial work based on Graph Convolutional Networks (GCN), which is one of the most powerful neural network architecture. GCN can properly find the similarity of the pixels and make a graph connection between them [22]. The proposed GCN model learns the features by inspecting neighboring nodes. The GCN network is used in the decoder network to construct accurate depth images in multi-scale as shown in Figure 2.

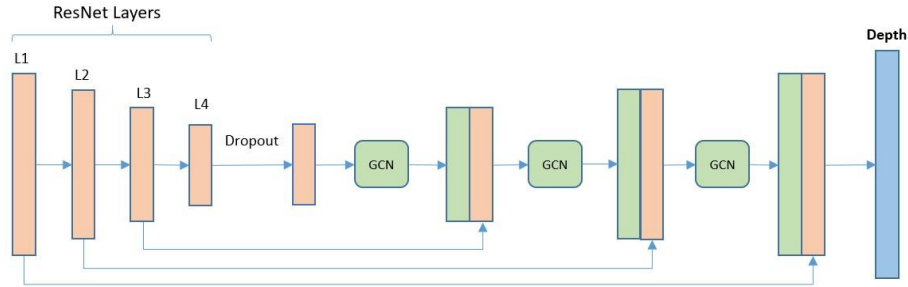

**Figure 2.** Overview of DepthNet network architecture

Regarding object detection, YOLOv5 is designed to create features from input images and later on feed these features through a prediction system to draw boxes around objects and predict their classes. This architecture consists of three main parts: Backbone, Neck and Head. YOLOv5 employed state-of-the-art network EfficientNet [23] as its backbone, making that the model has sufficientability to learn the complex features of input images. YOLOv5 applied an improved PANet[24], named bi-directional feature pyramid network (Bi-FPN) as its neck, to allow easy and fast multi-scale feature fusion. Bi-FPN introduces learnable weights, enabling the network to learn the importance of different input features, and repeatedly applies top-down and bottom-up multi-scale feature fusion. Thirdly, YOLOv5 integrates a compound scaling method that uniformly scales the resolution, depth, and width for all backbone, feature network, and box/class prediction networks at the same time, which ensures maximum accuracy and efficiency under the limited computing resources. Figure 3 demonstrates the architecture of YOLOv5 that we used to detect objects.

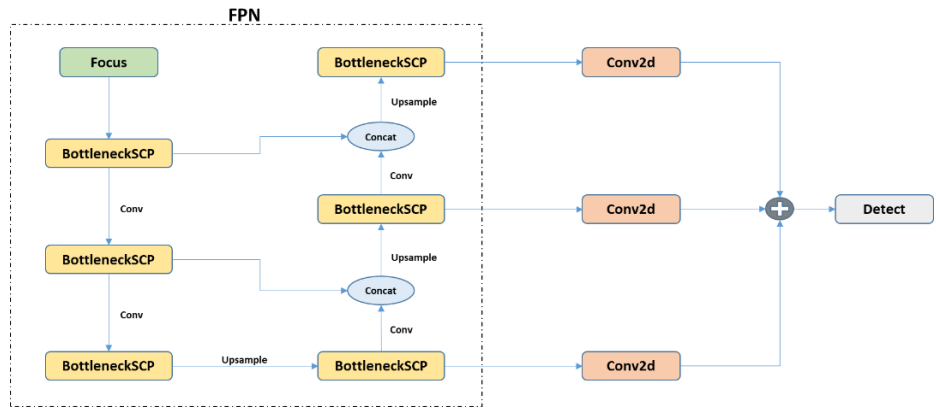

**Figure 3.** Overview of YOLOv5 network architecture.

*3.1. Absolute Distance Prediction*

After training our depth predicted model with a KITTI dataset [25], the model was tested with our private own dataset, explained in detail in section 4, to estimate the depth of each image in the testing set. At the same time, with the YOLOv5 model, objects in the image have been detected and the location of the bounding box was determined. Based on the bounding box coordinates, the exact box was localized on the corresponding predicted depth image. Indeed, the DepthNet estimates the disparity maps that represent the relation between the motion between the pixels of the input image and the ones of the target image (i.e., could be a consequence image). So, we transform the disparity maps to depth maps [26]. In the DepthNet network, minimum depth and maximum depth have been set as 0 to 100 meters same as the KITTI dataset.

Afterwards, the median value of the estimated distances of all pixels inside the bounding box of an object in a depth imageis computed. This estimated distance can be named the relative distance of an object (REV).

However to convert the REV distance to absolute distance (ABS), the real distance of objects in images are needed. Thenceforth, the relation between the absolute distance and relative distance have to be calculated. In most works of ABS estimation, ABS of an object depends on the type and shape of objects, as well as the image size and focal length of the sensor [27]. However, in this work, we need to avoid depending on this type of information and we will try to calibrate our method to work for different unknown objects.

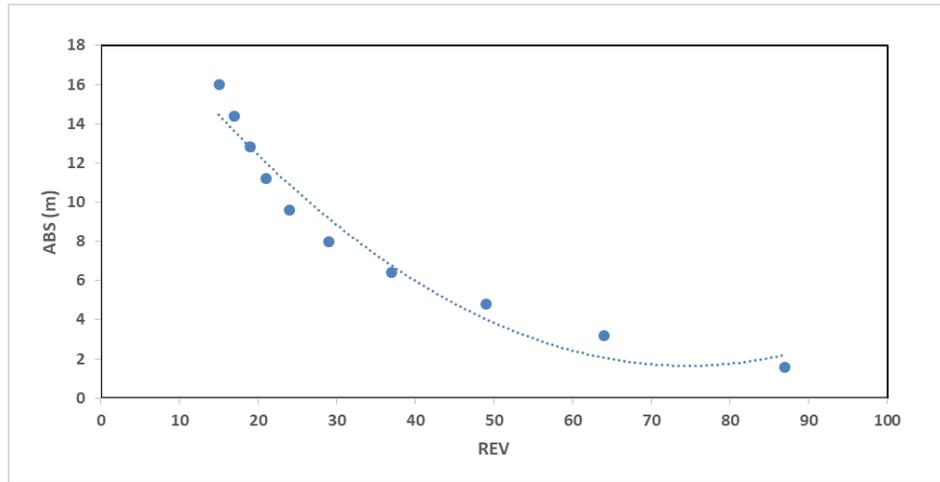

**Figure 4.**Perform the relation between ABS and REV

Consequently, based on the Taha and Jizat technique [28], the ABS distance can be calculated based on a mathematical quadratic function:

$$Y = (c_0 + c_1 X + c_2 X^2) \times h \qquad (1)$$

Where $c_0$, $c_1$, $c_2$ coefficients can be obtained using least square equations, $h$ is the camera height, and $X$ is the relevant distance from the object to the beginning of the camera's field of view. Based on this hypothesis, curve fitting and least-squared optimization is applied to find the approximated value of the four unknown coefficients. The solution has the best fit to a series of data points (i.e., we used 10 images with different objects and distances)to find the relation between ABS and REV distances, as shown in Figure 4:

$$Y = 0.0036X^2 - 0.5373X + 21.714 \qquad (2)$$

## 4. Experiments

*Implementation Details:* We implemented a code of depth estimation in Pytorch. The depth estimation model trained for 20 epochs, a batch size of 10, a learning rate of 0.0001 and the Adam optimizer. The training process took 5 days using a single GPU of GTX 1080 TI. For object detection with YOLOv5, we used also the Pytorch library with 80 different classes. Regarding the pre-trained checkpoints, YOLOv5 with a light version (YOLOv5s) has been chosen due to its lower computational cost.

*4.1. Datasets*

***KITTIdataset*** is one of the famous computer vision dataset for depth and pose estimation. The dataset contains 200 videos of street scenes in day light captured by RGB cameras and the depth maps captured by Velodyne laser scanner. We used synchronized single images from a monocular camera and Eigen split [29] with 39810 images for training, 4424 for validation and 697 images for testing. The [21] preprocessing method has been used for removing static frames. The resolution input size of the images is 320 pixels × 1024 pixels.

***Cocodataset*** [30] is one of the large-scale object detection, segmentation and captioning dataset. The dataset contains 80 different object classes with a total of 2.5 million labeled instances in 328k images. We used the original split dataset with 165482 images for training, 81208 for validation and 81434 images for testing. Theimages sizes are 640 pixels × 480 pixels.

From the predicted depths, the relative distance of each pixel can be extracted. However, having an absolute distance is necessary to first find the relation function and second evaluate the results. Therefore, we prepared our **own private dataset** that contains 100 images (with a resolution of 777 pixels × 1350 pixels) with a hand-held camera. Monocular RGB camera was mounted on a static stand and the absolute distance of each object was manually measured from the camera. The absolute distance from the camera and objects have manually been defined of all objects in scenes. During the collection of the dataset, we imitated potential static obstacles on the front of the camera. These obstacles were located in different distances from the camera test-stand.

*4.2. Evaluation*

During the test procedure, the performance of the proposed method was checked by using 100 images with different objects such as person, car, chair, etc. (i.e., remember we can detect 80 classes of objects as COCO used).

We used two standard evaluation measures to assess the proposed framework: Accuracy and Root Mean Square Error (RMSE). The Accuracy was used to estimate errors under a given threshold, serving as an indication of how often our estimation is correct. The threshold accuracy measure from [31] is essentially the expectation that the absolute distance value error of a given object in a scene is lower than a threshold T (i.e., in this work T is set by 0.2 m).

Figure 5 demonstrates the qualitative results of the proposed framework and it shows some examples of our own private dataset including the original images, estimated depth images, the results object localization using YOLOv5 and the relevant depth estimated by DepthNet. In turn, Table 1 represents the measurement of absolute distance vs. predicted distance. It is obvious from Figure 5, object detection based YOLOv5 is reliable in spite of the fact that YOLO classifier was used in its original form trained with COCO dataset, without fine-tuning with the images from own private dataset. Besides, it is obvious from Table 1 that achieved absolute distance estimation is satisfactory in spite of the fact that our own private dataset did not contain object boxes (object localization) from the captured scenes.

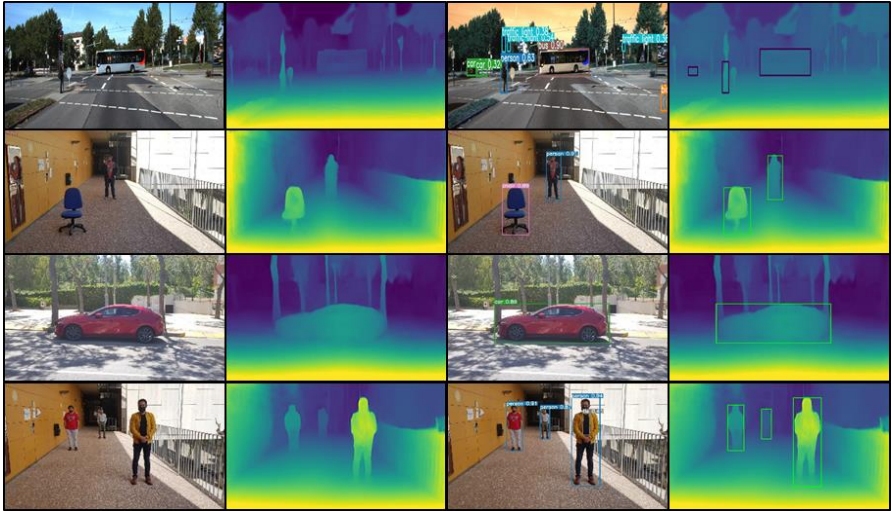

**Figure 5.** Visual process of whole network

**Table 1**. Estimated distance vs. absolute distance. Note the objects are counted in the tested images from left to right.

| Figure 5 | Object | Absolute distance (m) | Predicted distance (m) | Error (m) |
|---|---|---|---|---|
| Row 1 | Car | 53.9 | 53.21 | 0.69 |
| | Person | 21.5 | 21.35 | 0.15 |
| | Bus | 48.7 | 48.13 | 0.57 |
| Row 2 | Chair | 3.5 | 3.45 | 0.05 |
| | Person | 8.0 | 8.09 | 0.09 |
| Row 3 | Car | 10.1 | 9.83 | 0.27 |
| Row 4 | Person 1 | 8.0 | 8.13 | 0.13 |
| | Person 2 | 12.0 | 11.69 | 0.31 |
| | Person 3 | 4.0 | 3.88 | 0.12 |

As it is shown in Table 1, the farther away the objects are, the more error we will get. The proposed framework achieved an accuracy and average RMSE of 96% and 0.203 (m), respectively. It is obvious that the predicted distance is satisfactory despite the fact that our private dataset was not part of the training dataset for DepthNet and YOLOv5.

In contrast to the DispNet method [27], which detected only the objects on railways, the presented framework recognized different objects in the scene recorded by the own private dataset. As obvious, for big objects (e.g., cars), the YOLO network can easily detect them even with a large distance. However for small objects (e.g., chair or person), YOLO is sometimes not able to detect them from a large distance that degrade the performance of the proposed framework. Thus, in the future work, the YOLO will be updated to work with tiny objects.

## 5. Conclusion

In this paper, we proposed a reliable deep framework for estimating absolute distances of objects in real scenes. Our methods consist of 2 parallel networks; the first one is used to predict the depth values of images using a 2D monocular camera based on an unsupervised autoencoder network, and the second one detects the objects and extracts its localization box within the scene. Furthermore, the absolute distance of an object can be computed from relative distance by calibrating our framework with real images. Ongoing work, the absolute distance estimation will be achieved based on a learnable network for generalizing the framework for different objects and shapes and thus to improve the accuracy of distance estimation in own private dataset. Future work aims at developing an intelligent assistant system for aiding visual impaired people based on the proposed framework.

## Acknowledgments

This research has been possible with the support of the Secretariad' Universitatsi Recercadel Departamentd' Empresai Coneixement de la Generalitat de Catalunya (2020 FISDU 00405). We are thankfully acknowledging the use of the University of Rovira I Virgili (URV) facilities to carry out this work.